\documentclass[conference]{IEEEtran}
\IEEEoverridecommandlockouts
\usepackage{cite}
\usepackage{amsmath,amssymb,amsfonts}
\usepackage{algorithmic}
\usepackage{graphicx}
\usepackage{textcomp}
\usepackage{xcolor}
\usepackage{comment}
\usepackage{caption}
\usepackage{makecell} 
\usepackage{multirow}
\usepackage{flushend}
\usepackage{mathtools}
\usepackage{enumitem}
\usepackage{hyperref}

\usepackage{tikz} 
\usepackage{pifont}

\def\BibTeX{{\rm B\kern-.05em{\sc i\kern-.025em b}\kern-.08em
    T\kern-.1667em\lower.7ex\hbox{E}\kern-.125emX}}

\begin{document}
\bstctlcite{IEEEexample:BSTcontrol}

\title{
On Design Choices in Similarity-Preserving \\ Sparse Randomized Embeddings \\
\thanks{
The work of DK was supported by the European Union's Horizon 2020 Research and Innovation Programme under the Marie Skłodowska-Curie Individual Fellowship Grant Agreement 839179.
The work of DAR was supported in part by the Swedish Foundation for Strategic Research (SSF, grant nos. UKR22-0024 \& UKR24-0014) and the Swedish Research Council Scholars at Risk Sweden (VR SAR, grant no. GU 2022/1963). 
}
}

\author{
\IEEEauthorblockN{Denis Kleyko}
\IEEEauthorblockA{
\textit{Örebro University}, Örebro, Sweden \\
\textit{Research Institutes of Sweden}, Kista, Sweden \\
\href{mailto:denis.kleyko@oru.se}{denis.kleyko@oru.se}
}
\and
\IEEEauthorblockN{Dmitri A. Rachkovskij}
\IEEEauthorblockA{
\textit{Luleå University of Technology}, Luleå, Sweden\\
\textit{IRTC for IT and Systems}, Kiev, Ukraine\\ 
\href{mailto:dmitri.rachkovskij@ltu.se}{dmitri.rachkovskij@ltu.se},
\href{mailto:dar@infrm.kiev.ua}{dar@infrm.kiev.ua}
}
}

\maketitle

\begin{abstract}
Expand~\&~Sparsify is a principle that is observed in anatomically similar neural circuits found in the mushroom body (insects) and the cerebellum (mammals). 
Sensory data are projected randomly to much higher-dimensionality (expand part) where only few the most strongly excited neurons are activated (sparsify part). This principle has been leveraged to design a \texttt{FlyHash} algorithm that forms similarity-preserving sparse embeddings, which have been found useful for such tasks as novelty detection, pattern recognition, and similarity search. 
Despite its simplicity, \texttt{FlyHash} has a number of design choices to be set such as preprocessing of the input data, choice of sparsifying activation function, and formation of the random projection matrix. 
In this paper, we explore the effect of these choices on the performance of similarity search with \texttt{FlyHash} embeddings.
We find that the right combination of design choices can lead to drastic difference in the search performance.

\end{abstract}


\vspace{-0.6cm}
\section{Introduction}

Random projection (RP) is a well-known approach, which has been extensively studied in mathematics and computer science~\cite{JohnsonLindenstrauss1984, IndykApproximate1998, PapadimitriouLatent2000, VempalaRP2005}. 
It is also used in hyperdimensional computing/vector symbolic architectures framework~\cite{PlateHolographic1995, KanervaFully1997,  KleykoSurveyVSA2021Part1, GaylerJackendoff2003, KleykoSurveyVSA2021Part2} for formation of distributed representations of numeric vectors. RP leads to similarity-preserving randomized embeddings  that are useful for similarity search and classification in a host of applications in domains exemplified by computer vision, robotics, and natural language processing (see~\cite{xie2017survey} for an overview).
The use of RP has a rich history starting from the seminal result by Johnson and Lindenstrauss~\cite{JohnsonLindenstrauss1984}. In~\cite{IndykApproximate1998, VempalaRP2005}, linear embeddings using normally distributed components of the projection matrix were considered.
The results have been made even more practical by 
proving the same properties for RP with bipolar matrices (components from $\{-1, +1\}$) and (sparse) ternary matrices (components from $\{-1, 0, +1\}$)~\cite{AchlioptasFriendlyRP2003, LiSparseRP2006, KaneSparser2014}. 
In the context of hyperdimensional computing, RP with sparse ternary matrices was first applied in~\cite{KanervaRI2000, SahlgrenVector2001} and further developed and applied in~\cite{MisunoVector2005}. 
Further, in~\cite{charikar2002similarity}  dense binarization (by thresholding at zero) of the embeddings resulting from RP was considered. 
In~\cite{MisunoSearching2005, MisunoVector2005,  RachkovskijSimilarityRP2015, 
RachkovskijRP2012,
rachkovskij2015estimation}, it was proposed to use non-zero threshold(s) to binarize or ternarize the result of RP leading to sparse embeddings. 
Recently, in~\cite{DasguptaNeural2017} a neuro-inspired \texttt{FlyHash} algorithm to formation of sparse binary embeddings was proposed. 
In their essence, \texttt{FlyHash} relies on the use of  sparse binary RP matrices and sparsifying non-linear function, as suggested by Expand~\&~Sparsify principle that is observed in the fly olfactory~\cite{modi2020drosophila}.
Implementation-wise, \texttt{FlyHash} is a simple neural network where connections between the input and output layers represent the RP matrix, while the non-linear function is used for the activation of the neurons in the output layer.

In this paper, we focus mainly on several practical design choices that are available when forming neuro-inspired randomized embeddings with \texttt{FlyHash} and effects those choices can make on the performance of similarity search when using data from real applications. 
The design choices in focus are related to the following algorithmic aspects:

\setlist{nolistsep}
\begin{itemize}[noitemsep]
    \item Preprocessing of input data; 
    \item Distribution of values of the RP matrix; 
    \item Non-linear transformation following the RP.

\end{itemize}
Another aspect that has not been considered in the \texttt{FlyHash} study is the number of bits to represent the result of an embedding. The density of the randomized embeddings was only controlled from the point of view of amount of operations\footnote{
The computations needed for the sparsification were not considered in~\cite{DasguptaNeural2017}. 
We follow the same approach, 
leaving the computational aspects as a 
topic for future work.
} required to compute them. 
However, once the embeddings are computed they need to be stored in memory and, therefore, it is worth considering design choices available for optimizing the number of bits per embedding in addition to computations per embedding.  
To address this, we propose to use structured non-linearity producing block sparse codes and compare the memory requirements and performance of the resulting embeddings to those of the binary version of \texttt{FlyHash}. 
We demonstrate that when using a particularly favorable combination of the design choices, the results reported in the original \texttt{FlyHash} study~\cite{DasguptaNeural2017} 
can be improved significantly. 

The paper is structured as follows. 
Methods and materials that cover the considered design choices are described in Section~\ref{sect:methods}. 
The evaluation of the design choices is presented in Section~\ref{sect:evalution}. 
Section \ref{sect:discussion} provides discussion, touches on related work, and concludes with remarks for the future work.

\section{Materials and methods}
\label{sect:methods}

\subsection{Sparse randomized embeddings}
In this section, we describe the basic \texttt{FlyHash} algorithm for forming sparse randomized embeddings~\cite{DasguptaNeural2017} and some of its modifications that we position as design choices.

The input to the algorithm is a $d$-dimensional real-valued vector $\mathbf{x}$. 
The first step of the algorithm is an expansion of $\mathbf{x}$ to $D$-dimensional vector $\mathbf{y}$ using an RP matrix $\mathbf{M} \in [D,d]$ as: 
\noindent
\begin{equation}
    \mathbf{y}= \mathbf{M}\mathbf{x},
\label{eq:RP:res} 
\end{equation}
\noindent
where the RP matrix $\mathbf{M}$ is binary and sparse. 
Thus, if $M_{ij}$ is set to one this can be interpreted as the presence of connection from the $j$th component of $\mathbf{x}$ to the $i$th component of $\mathbf{y}$: 

\begin{equation}
    M_{ij}= 
\begin{cases}
    1,& \text{if $x_j$ }  \text{connects to } y_i\\
    0,              & \text{otherwise}
\end{cases}
\label{eq:RP:matrix} 
\end{equation}
\noindent

The second step is to form the vector $\mathbf{z}$ by sparsification of $\mathbf{y}$. 
A common way to do so is by defining a threshold value corresponding to the desired level of density in $\mathbf{z}$~\cite{KanervaSDM1988, RachkovskijRP2012}.
\texttt{FlyHash} uses a closely related idea of lateral inhibition that is formalized via $k$ Winner-take-all ($k$WTA) non-linear function that sets all but $k$ largest values of $\mathbf{y}$ to zero. 
Formally, values of components $z_i$ of $\mathrm{\mathit{k}WTA}(\mathbf{y})$ are computed as: 
\noindent
\begin{equation}
     z_{i}= 
\begin{cases}
    y_i,& \text{if $y_i$ is one of the $k$ largest entries in $\mathbf{y}$  }\\
    0,              & \text{otherwise}
\end{cases}
\label{eq:kWTA} 
\end{equation}
\noindent

One modification of the resulting embedding $\mathbf{z}$ in Eq.~(\ref{eq:kWTA}), 
is to neglect the actual values of $z_i$ by binarizing them~\cite{KanervaSDM1988,RachkovskijRP2012}. 
To obtain binary randomized embeddings, the components of $\mathbf{z}$ are further modified as: 
\noindent
\begin{equation}
     z _{i}= 
\begin{cases}
    1,& \text{if $z_i > 0$}\\
    0,              & \text{otherwise}
\end{cases}
\label{eq:kWTA:bin} 
\end{equation}
\noindent
Note that all binary $\{0,1\}$ vectors with $k$ 1-components have the same L2 norm $\sqrt{k}$. 
As an intermediate option between these two variants, we can perform Euclidean normalization (denoted as L2 or $||\cdot||_2$) of nonbinary $k$WTA:  $\frac{\mathbf{z}}{||\mathbf{z}||_2}$.
However, its performance is close to the binary $kWTA$, and so we do not report it in this study. Hereafter, if not stated otherwise, all normalizations are meant to be Euclidean.

While $k$WTA is a convenient mathematical abstraction, it might not be the best non-linear function when it comes to the engineering.
The components of $\mathbf{y}$ that will be kept active can be located anywhere in the embedding.
From the hardware implementation point of view, it is worth considering a non-linear function alternative to $k$WTA, that imposes a block constraint when constructing $\mathbf{z}$.
Such representations are known as block sparse codes~\cite{Laiho2015, GritsenkoAMSurvey2017, FradySDR2020}.
Similar to $k$WTA, in $k$-block sparse codes, the density of nonzero components is $k/D$.
The index set, is, however, partitioned into $k$ blocks where the size of each block is $D/k$ components and there is only one nonzero component per block:
\noindent
\begin{equation}
      z_{i}= 
\begin{cases}
    1,& \text{if $y_i$ is the largest component in a block}\\
    0,              & \text{otherwise}
\end{cases}
\label{eq:kBSC:bin} 
\end{equation}
\noindent
This block constraint can also be used to form nonbinary randomized embeddings, but we do not experiment with this variant in this study.

Naturally, the introduction of the block constraint reduces the information entropy (consequently, fewer states can be represented with it) from $\log_2\left( \binom{D}{k}\right)$ to $k \log_2\left(\frac{D}{k}\right)$ bits. 
At the same time, the number of storage bits necessary for block sparse codes is lower than that of $k$WTA representations. We investigate this aspect below in Section~\ref{sec:bsc:storage}.

\subsection{Datasets}

To demonstrate the effect of the design choices on the performance of $k$WTA embeddings, we use the same data as in~\cite{DasguptaNeural2017}: SIFT~\cite{JegouSIFT2011} ($d = 128$), GLOVE~\cite{PenningtonGlove2014} ($d = 300$), and MNIST~\cite{LecunGradient1998} ($d = 784$). 
Note that input vectors within SIFT and MNIST datasets have only non-negative components, whereas components of GLOVE take both positive and negative values.
In addition, all vectors in SIFT have very similar norms. 
For each dataset, a subset of $N = 10^4$ input vectors was selected.

\subsection{Performance metrics}

Following the evaluation protocol in~\cite{DasguptaNeural2017}, Mean Average Precision (MAP; see, e.g.,~\cite{LinHashing2013}) is used to assess the quality of preserving similarity by $k$WTA embeddings. 
MAP measures the overlap between the lists of true and predicted rankings of dataset's samples produced by the similarity measure of interest.
First, for each input vector $\mathbf{x}$ (query), the rankings of its (e.g., Euclidean) distances (or other similarity measure of interest) to all other vectors in a dataset is computed. Next, the second  list of rankings (using a similarity measure of interest) is computed with the corresponding $k$WTA embeddings. 
The Average Precision at $K$ is computed based on $K$ dataset vectors most similar to the query. 
We used $K=200$ as in~\cite{DasguptaNeural2017}.  
MAP at $K$ (below, we just say MAP) is obtained as the mean over the Average Precisions at $K$ across all the queries. 
To reduce the computational load, for each random realization of $\mathbf{M}$, we estimated the MAP from Average Precision at $K$ computed for $10^3$ queries randomly selected from the entire dataset. In the calibration experiments, this did not noticeably changed MAP compared to the values obtained from all $N=10^4$ queries.

\begin{figure*}[tb]
\centering
\includegraphics[width=1.93\columnwidth]{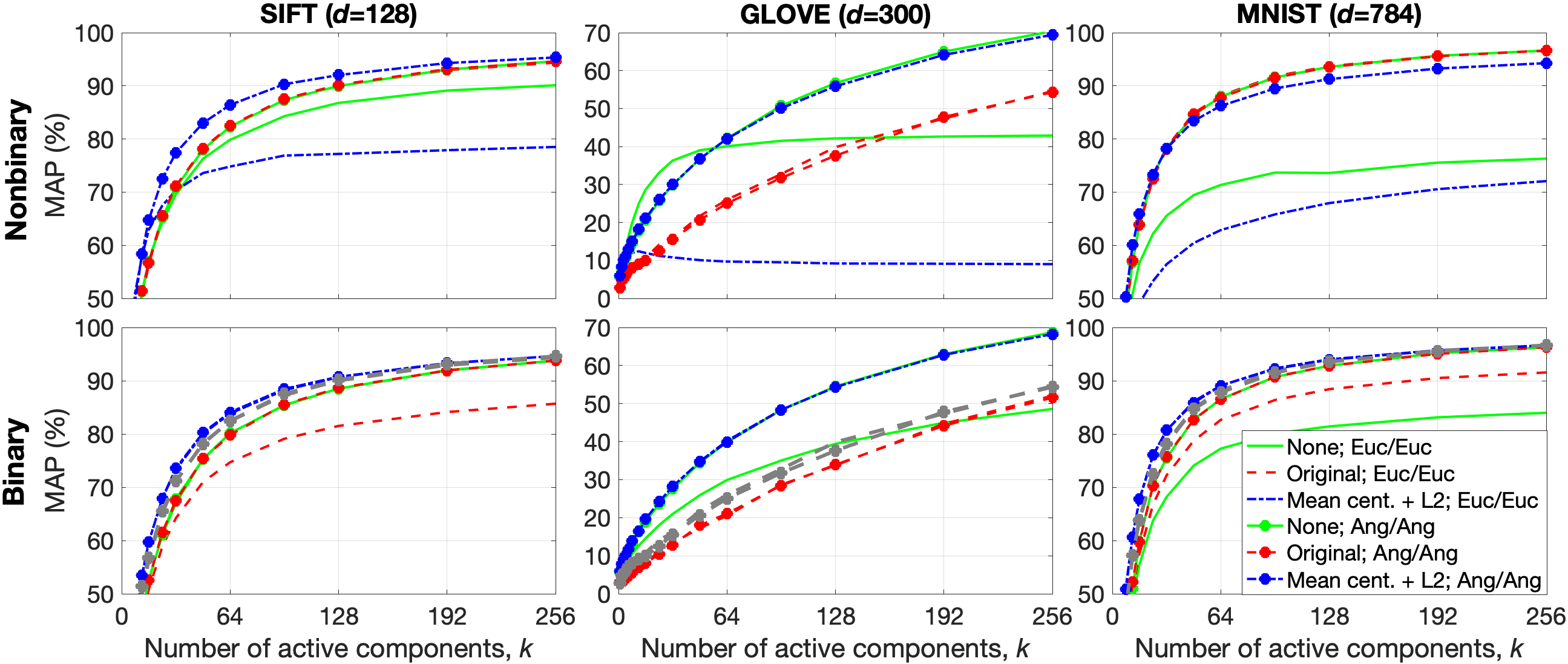}
\caption{
The effect of data preprocessing on the MAP obtained from the embeddings. 
Plots depict the dependency between MAP and the number of nonzero components $k$ in an embedding; $D$ was set to $20k$.  
Plots correspond to unique combinations of a dataset (columns) and an embedding variant (rows).
The values are averaged over $10$ random initializations of $\mathbf{M}$. 
\vspace{-0.6cm}
}
\label{fig:normalization}
\end{figure*}

\subsection{Data preprocessing techniques}
\label{sect:methods:preprocessing}

In \texttt{FlyHash} study, the preprocessing of data\footnote{
The description of the \texttt{FlyHash} preprocessing technique is based on the corresponding functions in the source code accompanying~\cite{DasguptaNeural2017}.
}
was done such that the components of each input vector $\mathbf{x}$ within a dataset are biased to become non-negative (the first step) and each input vector has the same mean (the second step).

Let us assume that the vectors of the dataset are available in matrix $\mathbf{X} \in [d,N]$, where $N$ is the number of vectors in the dataset. 
The first step of the original preprocessing technique was to increment each component of $\mathbf{x}$ with the absolute value of the minimum of this component in $\mathbf{X}$: 
\noindent
\begin{equation}
     \tilde{x}_i = x_i + |\min \mathbf{X}_{i:}  |,
\label{eq:norm:abs} 
\end{equation}
\noindent
where $\mathbf{X}_{i:}$ denotes the $i$th row.
Next, the mean value of $\tilde{\mathbf{x}}$ is rescaled to some predefined value $r$, which was set to $100$ 
in~\cite{DasguptaNeural2017}, and only the integer part of resulting values is kept: 
\noindent
\begin{equation}
     \widetilde{\mathbf{x}} = \left \lfloor \frac{r}{\frac{1}{d}\sum_{i=1}^d \tilde{x}_i } \tilde{\mathbf{x}}  \right \rfloor.
\label{eq:norm:mean} 
\end{equation}
\noindent
The result of Eq.~(\ref{eq:norm:mean}) is used in place of $\mathbf{x}$ to form its RP according to Eq.~(\ref{eq:RP:res}).

Given that there are numerous standard preprocessing techniques, we investigate how some of them will influence the MAP.
For example, an approach alternative to biasing each component in Eq.~(\ref{eq:norm:abs}) is 
``mean centering'' where the mean value of each component of $\mathbf{x}$ across the dataset is subtracted:
\noindent
\begin{equation}
     \widetilde{\mathbf{x}} = \mathbf{x} - \frac{1}{N} \sum_{j=1}^N \mathbf{X}_{:j},
\label{eq:norm:conc} 
\end{equation}
\noindent
where $\mathbf{X}_{:j}$ denotes the $j$th column.

Another common option alternative to preserving mean value of components as in Eq.~(\ref{eq:norm:mean}) would be to normalize all input vectors:
\noindent
\begin{equation}
     \widetilde{\mathbf{x}} =\frac{\mathbf{x}}{||\mathbf{x}||_2}.
\label{eq:norm:L2} 
\end{equation}
\noindent
Normalizing input vectors means that even if ranking the normalized input vectors is done by their Euclidean distances, the ranking is determined by the angles between vectors.

As yet another variant of preprocessing, we can combine together both mean centering \& normalization.
The preprocessed input vector is then computed as:
\begin{equation}
     \widetilde{\mathbf{x}} =\frac{\mathbf{x} - \frac{1}{N} \sum_{j=1}^N \mathbf{X}_{:j}}{||\mathbf{x} - \frac{1}{N} \sum_{j=1}^N \mathbf{X}_{:j}||_2}
\label{eq:norm:conc:L2} 
\end{equation}

Thus, the alternatives above provide us with at least four ways to preprocess input vectors prior to forming their embeddings. Below in Section~\ref{sect:evalution:preprocessing} we limit the reported results to the original and mean centering \& normalization preprocessings, as well as provide results for the input data without any preprocessing.

\begin{figure*}[tb]
\centering
\includegraphics[width=1.955\columnwidth]{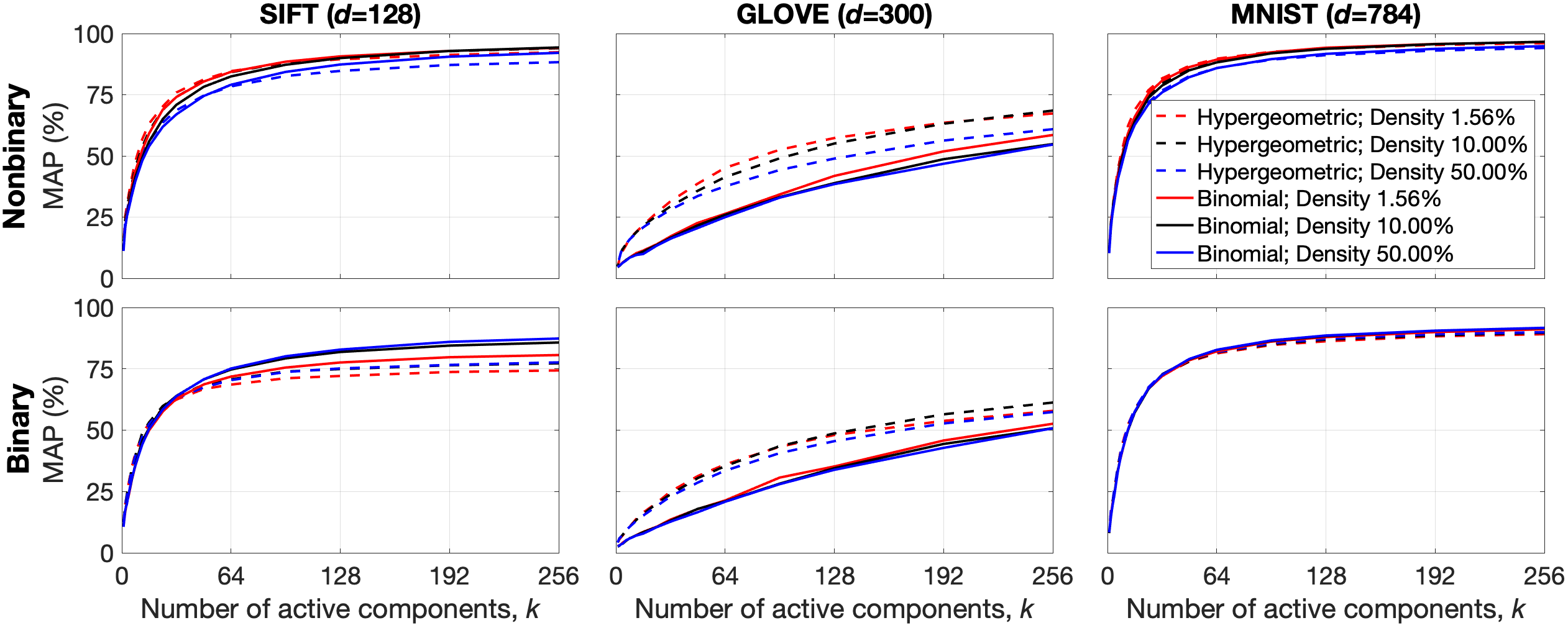}
\caption{
The effect of the distribution of the values in the RP matrix on the performance of the embedding. 
The plot depicts the dependency between MAP and the number of nonzero components $k$ in a randomized embedding; $D$ was set to be $20k$ (i.e., as $D$ grows, the density of $\mathbf{z}$ is fixed to $0.05$).  
The ranking of both input vectors and their embeddings was done by Euclidean distances.
Plots correspond to unique combinations of a dataset (columns) and an embedding variant (rows).
The datasets were processed with the original preprocessing, Eqs.~(\ref{eq:norm:abs})-(\ref{eq:norm:mean}).
The  MAP values are averaged over $10$ random initializations of $\mathbf{M}$. 
\vspace{-0.6cm}
}
\label{fig:distribution}
\end{figure*}

\section{Experimental results: role of design choices}
\label{sect:evalution}

\subsection{Effects of data preprocessing}
\label{sect:evalution:preprocessing}

Fig.~\ref{fig:normalization} presents the MAP for the preprocessing alternatives (Section~\ref{sect:methods:preprocessing}) against the number of nonzero components $k$ ($D=20k$) for different datasets (columns) and embedding variants (rows). 
Each plot depicts the results for two scenarios when both input vectors and embeddings were ranked either according to their angles (lines with markers, \texttt{Ang/Ang} ranking) or according to their Euclidean distances (lines without markers, \texttt{Euc/Euc} ranking). For (almost) all of the lines, we notice that MAP was increasing with $k$, that can be attributed to more accurate estimates of the similarity measures by higher-dimensional embeddings.
The noticeable exception to the above observation is mean centering \& normalization preprocessing (blue dash-dotted lines) combined with the nonbinary $k$WTA. For \texttt{Euc/Euc} ranking for all three datasets, this preprocessing resulted in the lowest MAP that is, especially, pronounced for GLOVE. The explanation to such a poor performance is that due to the normalization, ranking input vectors by Euclidean distances is implicitly ranking by angles. However, since norms of nonbinary $k$WTA embeddings are not guaranteed to be the same, \texttt{Euc/Euc} ranking is de facto two different similarity measures, leading to the poor performance. Note that, at the same time, once \texttt{Ang/Ang} ranking is considered, preprocessing by mean centering \& normalization became one of the best alternatives. 
Furthermore, \texttt{Ang/Ang} ranking also improved the performance substantially for the initial data (green lines), which is especially clear for GLOVE, where the corresponding line coincides with that of mean centering \& normalization preprocessing. For the original \texttt{FlyHash} preprocessing (red dashed lines), the results were very similar for both rankings. 
The binary $k$WTA (lower plots in Fig.~\ref{fig:normalization}) forms embeddings with exactly $k$ 1-components. So, their norms are the same, and, therefore, rankings by angles and Euclidean distances are identical. This explains why mean centering \& normalization  had the best performance across the datasets and showed identical results for \texttt{Ang/Ang} and \texttt{Euc/Euc} rankings. Furthermore, the results were higher than the performance of the original preprocessing with nonbinary $k$WTA (gray lines).
For the other preprocessing techniques, since they did not normalize input vectors (except for SIFT where initial norms are nearly the same), \texttt{Euc/Euc} ranking performed rather poorly (e.g., original preprocessing on SIFT). 
This experiment illustrates an important point that the performance on the target task (in our experiments, MAP as the measure of the quality of the similarity search) is strongly influenced by choosing the right data preprocessing,  proper variant of embedding (e.g., nonbinary versus binary $k$WTA as in our experiments), and selecting appropriate similarity measures for ranking (as exemplified in the above experiment by \texttt{Euc/Euc} and \texttt{Ang/Ang}).

\subsection{Effect of the distribution of the values in RP matrices}

\subsubsection{The choice of distribution and density of RP matrices}

In the \texttt{FlyHash} study, it has been suggested that the binary RP matrix $\mathbf{M}$ is sparse and can be formed as a realization of $Dd$ binomial experiments.
The use of the binomial distribution is intuitive, but from the design choice point of view it is interesting to consider the role of alternative probability distributions in the performance. A viable alternative to the binomial distribution is the hypergeometric distribution.  
Thus, if the density of nonzero components in $\mathbf{M}$ is $p$, when using the binomial distribution each component of $\mathbf{y}$ will on average get input from $s=pd$ components of $\mathbf{x}$. While in the case of the hypergeometric distribution, each component of $\mathbf{y}$ will get input from exactly $s=\lfloor pd \rceil$ components (i.e., rows of $\mathbf{M}$ are sampled from the hypergeometric distribution).\footnote{
We also considered the setup where each component of $\mathbf{x}$ was contributing to exactly $s=\lfloor pD \rceil$ components of $\mathbf{y}$ (i.e., columns of $\mathbf{M}$ were sampled from the hypergeometric distribution, as in~\cite{KaneSparser2014}. This did not provide any difference relative to the binomial distribution. A plausible explanation is that in~\cite{KaneSparser2014} the case of dimensionality reduction was considered, whereas we consider the case of dimensionality expansion. 
}
Therefore, the use of the hypergeometric distribution might be relevant in situations where $pd$ is small, as it guarantees the presence of input from some $\mathbf{x}_i$ to each component of $\mathbf{y}$. 
For the considered expansion case $D > d$  it implies that with a high chance every component of $\mathbf{x}$ will contribute to some component(s) of $\mathbf{y}$, even when $pd$ is small.

\begin{figure*}[tb]
\centering
\includegraphics[width=1.95\columnwidth]{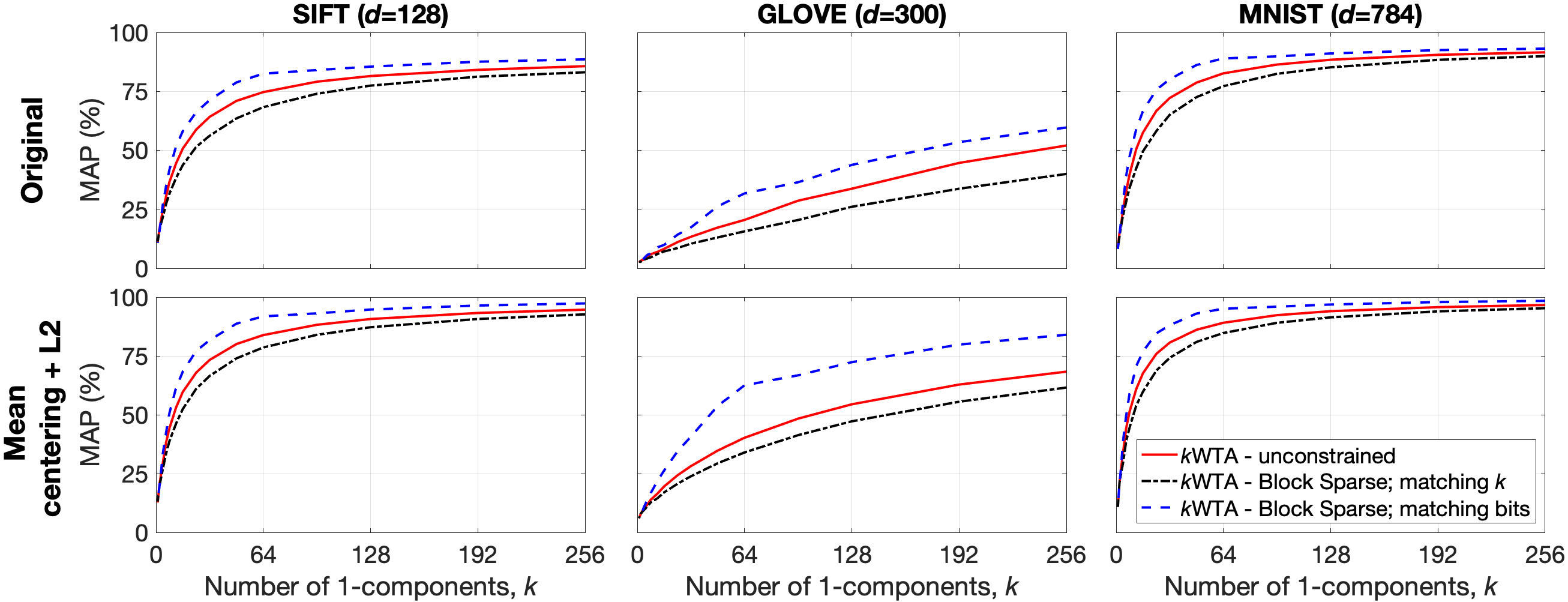}
\caption{
The effect of the block sparse non-linearity on the performance of the embeddings. 
Each plot depicts MAP against the number of 1-components $k$ in an embedding; $D$ was set to be $20k$.  
The ranking of both input vectors and their embeddings was done by Euclidean distances.
Plots correspond to unique combinations of a dataset (columns) and type of preprocessing (rows).
The reported values are averaged over $10$ random initializations of $\mathbf{M}$. 
\vspace{-0.6cm}
}
\label{fig:sparse:block}
\end{figure*}

When it comes to the density of nonzero components in $\mathbf{M}$, in~\cite{DasguptaNeural2017} $p$ was set to $0.1$, i.e., on average $10.0\%$ of components of $\mathbf{M}$ are set to one. 
Here, in addition to the choice of distribution of $\mathbf{M}$ (binomial versus hypergeometric), we use three levels of density of $\mathbf{M}$: very sparse ($1.56\%$), sparse ($10.0\%$), and dense ($50.0\%$) to explore its role on the performance. 
Fig.~\ref{fig:distribution} presents MAP for six unique combinations of the distribution and density against the number of nonzero components $k$ in $\mathbf{z}$ for different datasets (columns) and embedding variants (rows).

The results were obtained using the original preprocessing, Eqs.~(\ref{eq:norm:abs})-(\ref{eq:norm:mean}) in the \texttt{Euc/Euc} scenario.
The observed results depended primarily on the dataset. 
For both variants of $k$WTA, there was no substantial difference in the performance when decreasing the density of ones even to the very sparse level (SIFT and MNIST).
There was, however, large difference in the case of GLOVE where the performance increased substantially for the hypergeometric distribution for both the nonbinary (Eq.~(\ref{eq:kWTA}); upper plots) and binary (Eq.~(\ref{eq:kWTA:bin}); lower plots) $k$WTA. The main difference leading to this observation is the density of nonzero components in the input vectors, since data in MNIST and SIFT are rather sparse while GLOVE is fully dense.

\subsubsection{Varying density of columns of \textbf{M}}

For the results reported in Fig.~\ref{fig:distribution}, we have fixed the density of ones in the binomial RP matrix $\mathbf{M}$ to a single value (so that we expect the same mean number of ones in each column). 
This assumption is in line with the original algorithm. 
However, it is known, that in the fly olfactory the connectivity between projection neurons (input layer) to Kenyon cells (output layer) that corresponds to $\mathbf{M}$ can demonstrate varying density of components in different columns of $\mathbf{M}$~(cf. Fig.~3 in \cite{CaronRandomOlfactory2013}).
Therefore, there is a question on whether the density of individual columns of $\mathbf{M}$ can be chosen such that the performance of $k$WTA embeddings is improved.

We used the RP matrix $\mathbf{M}$ drawn with the binomial distribution and binary $k$WTA embeddings together with the genetic algorithm that was set to optimize the density of individual columns of $\mathbf{M}$ with respect to the performance of the embeddings. 
We have not found any noticeable improvements once the optimization process was done.

\subsection{Effect of block sparse codes}

\subsubsection{Effect of considering the number of bits per embedding}
\label{sec:bsc:storage}

In~\cite{DasguptaNeural2017}, the focus of comparison studies was on  the amount of computations required to form $k$WTA embeddings.
There is, however, another dimension to $k$WTA embeddings -- that is, the number of bits to represent them.
For the sake of simplicity, let us only consider binary $k$WTA embeddings formed, e.g., via the binary $k$WTA.
Since the resulting embeddings are sparse, there is only need to represent the locations of  $k$ 1-components chosen after $k$WTA.

As a 1-component can be located anywhere in a $D$-dimensional embedding, a single location can be represented with $\log_2(D)$ bits, and then the cost (in bits) of representing the whole embedding is: 
\noindent
\begin{equation}
     k\log_2(D).
\label{eq:bin:mem:kWTA} 
\end{equation}
\noindent
In the case when randomized embeddings' 1-components are chosen using the block constraint, Eq.~(\ref{eq:kBSC:bin}), one only needs to represent a location of a 1-component within its corresponding block, since it is known that there is strictly one 1-component per each of the $k$ blocks.
Thus, it only costs $\log_2(D/k)$ bits and the whole embedding then requires (in bits): 
\noindent
\begin{equation}
     k\log_2(D/k).
\label{eq:bin:mem:block} 
\end{equation}
\noindent
Thus, one option when using $k$-block sparse codes is to form embeddings with the total number of bits lower than that for binary $k$WTA at the same $D$ and $k$.

\begin{figure*}[tb]
\centering
\includegraphics[width=1.95\columnwidth]{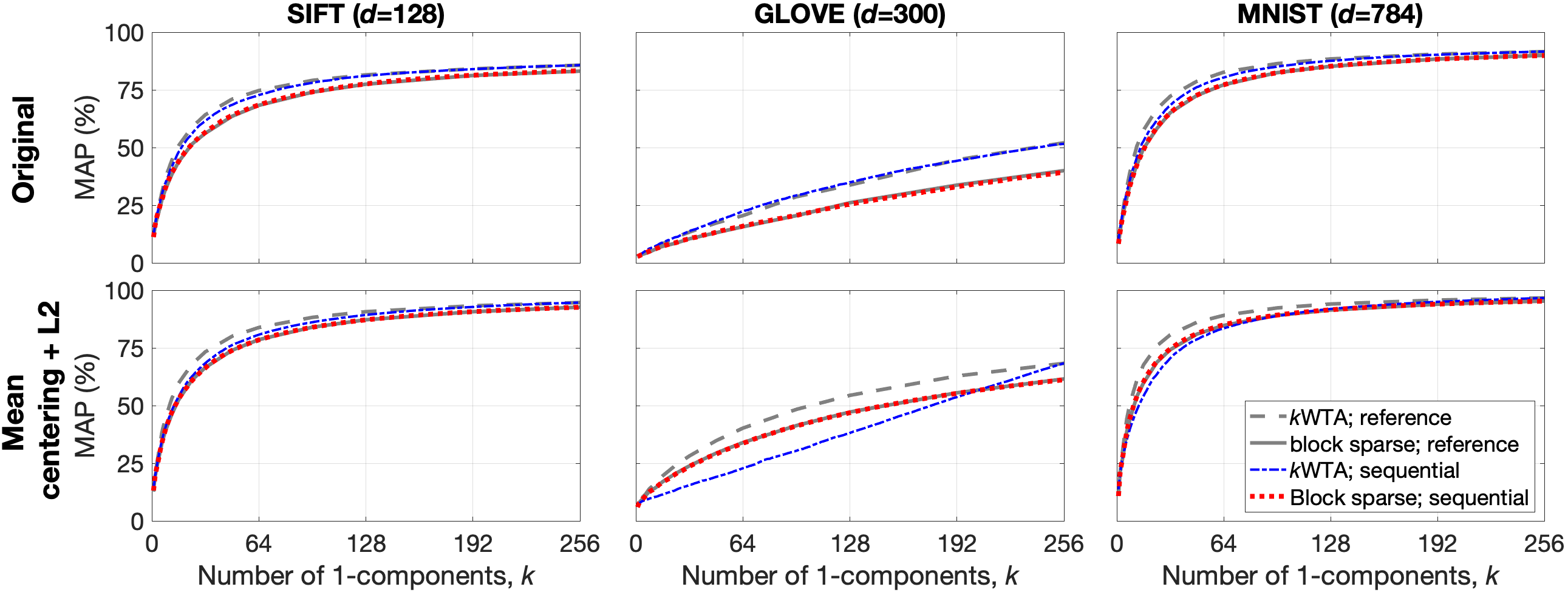}
\caption{
The effect of the ``sequential'' processing of randomized embeddings (dash-dotted and dotted lines).
Gray lines (solid and dashed) are the corresponding results from Fig.~\ref{fig:sparse:block} and used as the reference.
A plot depicts the dependency between MAP and the number of processed blocks of a randomized embedding with $k=256$; the block size was fixed to $20$ so $D$ was  set to be $5,120$.
The ranking between both input vectors and their embeddings was by Euclidean distances.
Plots correspond to unique combinations of a dataset (columns) and type of preprocessing (rows).
The reported values are averaged over $10$ random initializations of $\mathbf{M}$. 
\vspace{-0.6cm}
}
\label{fig:sparse:block:seq}
\end{figure*}

Another alternative is to use block sparse codes with the same number of representing bits as in $k$WTA embeddings by increasing the number $k'$ of 1-components that can be computed for any given $k$ and $D$ by solving 
\noindent
\begin{equation}
     k'\log_2(D/k')=k\log_2(D)
\label{eq:bin:mem:block:mat} 
\end{equation}
\noindent
for $k'$.
Practically, this will result in randomized embeddings with the total number of bits that is less than or equal to that of the randomized embeddings formed with the binary $k$WTA, but with the higher density of  1-components as $k'>k$.

In order to assess the effect of the usage of block sparse non-linearity on the performance, we performed the experiments reported in Fig.~\ref{fig:sparse:block}.
The figure depicts the results for the binary $k$WTA (solid red lines) and for two variants of block sparse codes. 
The variant denoted as ``block sparse; matching $k$'' (dash-dotted black lines) corresponds to the case when the number of blocks is the number of nonzero components (i.e., $k$, as for embeddings formed with the binary $k$WTA). 
The second variant denoted as ``block sparse; matching bits'' (dashed blue lines) used $k'$ blocks matching the number of bits to represent embeddings formed with the binary $k$WTA as in Eq.~(\ref{eq:bin:mem:block:mat}). 
The experiments were performed for two types of preprocessing: original preprocessing (upper plots) and mean centering \& normalization (lower plots).

The obtained results are consistent across datasets and preprocessings. 
First, when replacing $k$WTA with $k$-block sparse codes, the performance of the block sparse codes was worse.
We attribute this to the block constraint, which forces some of the $k$ largest components of RP's result $\mathbf{y}$  to be set to zero, while setting to one some components that do not belong to the $k$ largest ones. 
However, for all plots, the second variant (``matching bits'') of block sparse codes consistently outperformed two other alternatives and demonstrated the highest performance.
This is an important observation as it suggests that the block sparse codes provide higher performance per bit than the binary $k$WTA.

\subsubsection{Effect of sequential processing}

Here we investigate how the performance changes if an embedding is processed in a ``sequential'' manner -- one block after the other, both for $k$-block sparse codes and for the binary $k$WTA. 
To do so, we compute MAP for the dimensionality of embeddings $D'=iD/k$, where $i$ is the number of blocks being combined. 
In fact, for the case of $k$WTA we could expect some difference in the performance, since in the absence of the block constraint there would be embeddings with some of blocks without 1-components. Hence, the processing of such blocks would not improve the performance.

The results are reported in Fig.~\ref{fig:sparse:block:seq}. 
We performed the experiments for randomized embeddings with the block size of $20$ and $k=256$ (so $D=5120$)  and measured the MAP after incrementally processing $i \in [1,k]$ consecutive blocks ($x$-axis). 
This was done for both embeddings formed with the binary $k$WTA, Eq.~(\ref{eq:kWTA:bin}), and with $k$-block sparse codes, Eq.~(\ref{eq:kBSC:bin}).
The most important observation is that  the mean performance after sequentially processing $i$ blocks (dotted red lines) was the same as the performance of block sparse codes with exactly $i$ 1-components (solid gray lines) that was reported in Fig.~\ref{fig:sparse:block}. 
This was not always the case for binary $k$WTA embeddings (dash-dotted blue lines).
In fact, for many configurations (cf., e.g., lower-right plot) the corresponding $i$WTA embedding (dashed gray lines) resulted in worse performance. Expectedly, once all the blocks were processed, the two lines would meet.

\section{Discussion}
\label{sect:discussion}

\subsubsection*{\textbf{Expand~\&~Sparsify principle}} In this study, we focused on similarity-preserving sparse randomized embeddings obtained from the \texttt{FlyHash} algorithm that is based on the Expand~\&~Sparsify principle observed in different brain areas across organisms as diverse as humans, fruit flies, and electric fish~\cite{modi2020drosophila}.  
Within neural networks, one of the earliest encounters of leveraging the principle was to use it as an address mechanism in high-capacity associative memories~\cite{KanervaSDM1988}. 
Recently, the principle has been found valuable for various areas within machine learning such as designing binary autoencoders~\cite{osaulenko2021expansion}, learning word embeddings~\cite{liang2021flyembeddings}, analyzing the attention mechanism~\cite{bricken2021attention}, and designing mechanisms to tackle the catastrophic forgetting~\cite{shen2023reducing, trauble2023discrete, bricken2023sparse}.  
This wide range calls for investigating how the design choices available within \texttt{FlyHash} affect the resulting embeddings. 
We focused on three concrete aspects:   
preprocessing of input vectors, sampling of random projection matrices, and the choice of non-linear transformation leading to sparse embeddings.

\subsubsection*{\textbf{Preprocessing techniques and non-linearity}} Following its neural inspiration, the \texttt{FlyHash} study proposed a particular preprocessing,~Eq.~(\ref{eq:norm:abs})-(\ref{eq:norm:mean}) that can be treated as the frequency of neurons' activations. However, other standard preprocessing techniques can be more beneficial in applications. 

There are theoretical results in the spirit of approximation of kernel methods~\cite{rahimi2007random, fradyfunctionsnice2022}, which concern input data transformations that allow for preservation of various similarity measures. 
As for the linear random projection, generally it preserves 
the Euclidean distances, inner products, and angles  between the input vectors when the same measure is computed on the embeddings. 
A seminal result of a non-linear transformation is the random projection followed by the binarization with zero thresholding~\cite{charikar2002similarity} that allows estimating the angle between the real-valued input vectors.
In the case when the threshold is set above zero, similarity measures on the resulting sparse binary embeddings are monotonic functions of the angles or cosine similarity between the input vectors.  
The behavior of $k$WTA is evidently close to thresholding that returns embeddings with $k$ ones (on average), especially for large values of $D$. This brings up the point that the similarity-preserving properties of various transformations should be considered and exploited. For example, the embeddings from the binary $k$WTA have the same norms, so their ranking by both Euclidean distances and angles is identical. Therefore, as illustrated in Fig.~\ref{fig:normalization} (e.g., original preprocessing in lower plots), unless input vectors have the same norms, the ranking of their Euclidean distances will not be preserved well by the binary $k$WTA embeddings. So, the input vectors need to be either ranked explicitly using their angles or prepossessed to have the same norm before measuring the Euclidean distance, as was the case for mean centering \& normalization in Section~\ref{sect:evalution:preprocessing}.
      
Note that in Fig.~\ref{fig:normalization}, we compared two variants of $k$WTA based on the number of active elements $k$ in the embeddings.  
However, embedding to the binary $k$WTA requires much fewer bits to be represented, Eq.~(\ref{eq:bin:mem:kWTA}), compared to the nonbinary $k$WTA, where the nonzero components are real-valued and, thus, would require $n$ times more bits, where $n$ is the number of bits for the chosen precision (e.g., $32$). Therefore, in future work we are planning to explore the performance of the binary and nonbinary $k$WTA, while changing the density of the binary embeddings relative to the real-valued embeddings. An additional practical consideration in favor of binary embeddings is that they are highly amenable for processing on specialized digital hardware~\cite{thomas2022streaming}.

\subsubsection*{\textbf{Sampling of projection matrices}}
As per the expansion step, we have focused on the case when it is done using the binary random projection matrix sampled from either the binomial or hypergeometric distributions. 
We noted that the effect of the choice of distribution was much more pronounced compared to the density of 1-components in the random projection matrix. 
Moreover, the effect from the hypergeometric distribution was evidently connected to the density of input vectors -- as it mostly affected GLOVE (fully dense) and barely changed the performance on MNIST (density 19\%).
There are other relevant ways of constructing projection matrices such as in~\cite{KaneSparser2014}, that should be considered in the follow-up investigations. 
Furthermore, there are relevant results in~\cite{ryali2020bioLSH, bricken2023sparse, pourmand2024laplace} demonstrating that better performance can be achieved when the projection matrix is learned to adapt to the input data. 
In fact, it was shown in~\cite{shen2023reducing, trauble2023discrete} that it can be sufficient to perform the expansion on a subset of randomly chosen samples from the dataset. 
This step will likely reduce the computational efficiency of the expansion step since the projection matrix will only be sparse if the dataset is sparse. 
Therefore, a direction for future work is to investigate ways of generating sparse projection matrices that are still adapted to the input data.

\subsubsection*{\textbf{Block sparse codes}} As a  specific way of realizing $k$WTA, we considered $k$-block sparse codes. As noted above, due to the block constraint, fewer states can be represented with such codes compared to the unconstrained binary $k$WTA. Therefore, worse performance that we observed in Fig.~\ref{fig:sparse:block} was rather anticipated. 
Similar problems were observed when using block sparse codes within associative memories. It was, however, shown in~\cite{gripon2011,KnoblauchPalm2019} that the block constraint can be used to improve the retrieval of stored patterns from the associative memory, so that the number of stored patterns in such memories is on a par with unconstrained $k$WTA. 
We leveraged the block constraint from the point of view of efficiently storing 1-components of the embeddings (cf. Eq.~(\ref{eq:bin:mem:kWTA}) \& Eq.~(\ref{eq:bin:mem:block})) by increasing the density of 1-components relative to the unconstrained binary $k$WTA. This led to better performance (cf. Fig.~\ref{fig:sparse:block}) for different preprocessing techniques. In future work, it will be important to assess the change in the performance when instead of increasing the number of blocks $k'$ (so increasing the density of 1-components) the block size is expanded (so decreasing the density of 1-components). 
The other practical advantage of $k$-block sparse codes is the ability of sequential processing (cf. Fig.~\ref{fig:sparse:block:seq}) as well as the simplicity of computing the similarity between two embeddings. If represented by the positions of their 1-components, only the matches between the positions within the same blocks needs to be considered. Given these advantages of $k$-block sparse codes, it will be important to better link them to the already existing theoretical analysis of $k$WTA embeddings~\cite{dasgupta2020expressivity}.

\subsubsection*{\textbf{Final remarks}}

A particular limitation of this study is that we only focused on studying the preservation of rankings between the vectors in a dataset, which implies that the focus was on investigating how embeddings preserve the similarities between the corresponding input vectors. Furthermore, as it was shown in Section~\ref{sect:evalution:preprocessing}, the choice of preprocessing can substantially affect the performance due to its effect on the similarity measure. More generally, it is important that the embeddings preserve  the particular similarity measure beneficial to the downstream task.
For example, expanding by the linear random projection is not meaningful if the goal is to only estimate a similarity measure of input vectors by their embeddings, as it cannot be preserved better than the original vectors do.
However, the linear projection may be useful in some other cases, such as increasing dimensionality to increase the information capacity of randomized high-dimensional representations~\cite{FradyCapacity2018, SchlegelVSAComparison2020,hersche2021near,thomas2021theoretical,clarkson2023capacity, kleyko2023efficient}, so that one can represent more complex object descriptions in such representations~\cite{NeubertAggregation2021,trauble2023discrete}.
On the other hand, adding a non-linear function on top of the linear random projection can change the estimated similarity measure in a manner useful for a particular application, or extracts non-linear features from the input vectors. 
Such non-linear features can be very useful when the goal is to solve a classification problem.
Then, following the spirit of kernel methods, the embeddings can be combined with linear models to classify data with non-linear class regions.
Given the substantial interest within hyperdimensional computing for solving classification problems~\cite{KleykoSDR2016, GeClassificationReview2020}, investigating the performance of \texttt{FlyHash} embeddings on benchmarks (e.g.,~\cite{Dua2019}) could be another fruitful direction in the future.

\section{Acknowledgments}
DK thanks Saket Navlakha and Sanjoy Dasgupta for providing the access to the original implementation of \texttt{FlyHash} and the related experiments as well as for stimulating discussions on the early ideas that led to this study.

\bibliographystyle{IEEEtran} 
\bibliography{References}

\end{document}